\documentclass[10pt,conference,a4paper]{IEEEtran}

\usepackage{times}
\usepackage{soul}
\usepackage{epsfig}
\usepackage{url}
\usepackage[hidelinks]{hyperref}
\usepackage[small]{caption}
\usepackage{graphicx}
\usepackage{amsmath}
\usepackage{booktabs}
\usepackage{makecell}
\usepackage{amssymb}
\usepackage{color, colortbl}
\definecolor{LightCyan}{rgb}{0.88,1,1}
\usepackage[shortlabels, inline]{enumitem}
\usepackage[ruled,vlined]{algorithm2e}
\ifCLASSINFOpdf
\else
\fi
\def\ie{\emph{i.e}.}

\begin{document}
%
\title{Pose-based Body Language Recognition for Emotion and Psychiatric Symptom Interpretation}

\author{\IEEEauthorblockN{Zhengyuan~Yang$^1$, Amanda~Kay$^2$, Yuncheng~Li$^3$, Wendi~Cross$^2$, and Jiebo~Luo$^1$}
\IEEEauthorblockA{$^1$Department of Computer Science, University of Rochester}
\IEEEauthorblockA{$^2$Department of Psychiatry, University of Rochester Medical Center}
\IEEEauthorblockA{$^3$Google Inc.}
}


\maketitle

\begin{abstract}
Inspired by the human ability to infer emotions from body language, we propose an automated framework for body language based emotion recognition starting from regular RGB videos. In collaboration with psychologists, we further extend the framework for psychiatric symptom prediction. 
Because a specific application domain of the proposed framework may only supply a limited amount of data, the framework is designed to work on a small training set and possess a good transferability. The proposed system in the first stage generates sequences of body language predictions based on human poses estimated from input videos. In the second stage, the predicted sequences are fed into a temporal network for emotion interpretation and psychiatric symptom prediction. We first validate the accuracy and transferability of the proposed body language recognition method on several public action recognition datasets. We then evaluate the framework on a proposed URMC dataset, which consists of conversations between a standardized patient and a behavioral health professional, along with expert annotations of body language, emotions, and potential psychiatric symptoms. The proposed framework outperforms other methods on the URMC dataset.
\end{abstract}

\IEEEpeerreviewmaketitle


\section{Introduction}
\label{sec:introduction}
Humans have shown a remarkable ability to infer emotions in face-to-face conversations, and much of the inference is made through body language. For example, among people in similar culture, ``touching one's nose'' implies disbelief, and ``holding one's head in the hands'' expresses upset. It seems a natural ability for humans to understand the ``meaning'' of body language. To help machines acquire a similar ability, we propose a two-stage framework that predicts emotions based on body language with regular RGB video inputs. 
In the first stage, the model predicts body language from input videos based on estimated human poses. The predicted body language are then fed into the second stage for emotion interpretation. We define a body language as a certain maintained posture or a period with repeated short actions. It is similar to but different from human actions defined in previous studies~\cite{kuehne2011hmdb,soomro2012ucf101}, which are usually shorter in time and contain dynamic motions. For example, ``holding one's head in the hands'' is a typical body language and is informative for emotion recognition, but it is not an action according to the definition in action recognition~\cite{kuehne2011hmdb,soomro2012ucf101}.

Automated body language based emotion recognition is useful in various application domains, such as health care, online chat, and computer-mediated communication~\cite{beyan2017moving}. Despite the shared automated body language and emotion recognition techniques in different application domains, the body language and emotion of interest vary. For example, online chatting systems are concerned with detecting people's moods, \ie, whether they are happy or not, while applications in health-care scenarios focus on identifying potential signs of mental disorders such as depression or panic attacks. Since a specific emotion can only be reflected by the corresponding body language\footnote{Examples of body language and their meanings can be found in this website. https://www.enkiverywell.com/body-language-examples.html}, different applications require the annotation of different body language and emotions. Annotating videos for each application potentially lead to a high annotation cost. To alleviate the data annotation problem, we design our framework to learn from a small training set together with the expert knowledge, instead of being purely data-driven. 

Specifically, we improve the method's transferability and reduce the required amount of annotations by 1). using the abstracted human poses as the framework input, 2). proposing a KNN based approach for body language recognition, and 3). conducting emotion recognition purely based on the predicted body language sequences.
In the first-stage body language recognition, instead of directly predicting body language from RGB videos~\cite{simonyan2014two,tran2015learning}, we use human poses as the input for body language recognition. Human poses are groups of human joint coordinates that provide abstracted high-level human structural information. Because of the abstractness, pose-based methods require less training data and have better transferrabilities, as proved empirically in our study. The recently avabile robust pose estimation methods~\cite{cao2016realtime,simon2017hand,wei2016cpm} make our pose-based attempt more feasible. We adopt Openpose~\cite{cao2016realtime} for pose estimation, and propose a Spatio-Temporal Convolutional Pose Feature (ST-ConvPose) to encode the spatial-temporal relationships among the joints. With the learned pose representation, we propose a K-Nearest-Neighbors-based classifier for body language recognition. In the second-stage emotion recognition, the model predicts the emotion based on the predicted body language sequences. 

Furthermore, as a concrete example of application, we adopt the proposed framework to help psychiatrists understand psychiatric symptoms from patients' body language and emotions. In collaboration with psychologists, we collected the URMC dataset under the mental healthcare scenario. The recorded videos are the conversations between a standardized patient and a psychiatrist or psychologist. Health experts define the body language, emotion, and psychiatric symptom classes adopted in this study. Multiple psychologists annotate the recorded videos in the URMC dataset with the defined classes. Experiments on the URMC dataset prove the effectiveness of the proposed framework.
%

Our main contributions are two-fold:
\begin{itemize}
\item We propose a framework to infer emotions from the body language. We design the framework to be interpretable and transferrable. 
\item We adopt the proposed framework to help mental health professionals predict psychiatric symptoms. A new URMC dataset is built for the study.
\end{itemize}

\section{Related Work}
Human action recognition from videos is an important area in computer vision. 
Most of the RGB video based action recognition studies~\cite{simonyan2014two,donahue2015long,tran2015learning} start from low-level features. Intuitively though, high-level features such as human joints should be informative and beneficial for boosting the recognition accuracy. Joint-based action recognition~\cite{ke2017new,liu2016spatio,yan2018spatial,yang2018actionicpr,yang2018action} 
acquires reliable 3D joints with Kinect or similar RGB-D sensors and generates action predictions based on the joint sequences. However, one inherent problem in applications is the cost and inconvenience of depth sensors. Furthermore, depth sensors are difficult to integrate into the application system.

To better capture human-related features, Jhuang et al.~\cite{jhuang2013towards} propose the scheme of using both RGB videos and poses for action recognition. Pose-CNN~\cite{cheron2015p} proposes to use joint information to generate body parts sub-images, which are fed into a two-stream network for action recognition. Studies in JHMDB~\cite{jhuang2013towards} and Pose-CNN~\cite{cheron2015p} suggest that high-level information is rewarding for action recognition, while subject to the limited performance of pose estimation. The recent improvements in human pose estimation~\cite{cao2016realtime,papandreou2017towards} make pose-based recognition feasible and attractive. Several recent studies~\cite{cao2016action,zolfaghari2017chained,iqbal2017pose,du2017rpan} propose various further improvements on the RGB+Pose task. We conduct body language prediction from RGB videos with poses calculated by OpenPose~\cite{cao2016realtime,simon2017hand}. 

This study is also related to 
emotion recognition~\cite{mckeown2012semaine,busso2008iemocap,martin2006enterface,you2016building, rao2016learning,han2017hard,zhao2017learning,chen2016emotion,machajdik2010affective,yang2019human,kleinsmith2012affective,karg2013body,de2005towards,kleinsmith2005grounding}. 
Several previous studies~\cite{lin2014user,sidorov2014emotion} propose to detect psychological stresses with multi-modal contents. Xu et al.~\cite{xu2014temporal} recognize effects with body movements.
Furthermore, this paper shares a similar goal with several previous studies on predicting human emotion from body motions and facial expressions. Traditional studies~\cite{gong2007visual,tang2013deep} focus on analyzing facial expressions for emotion interpretation. De et al.~\cite{de2006towards} first propose to adopt gestures as a channel for emotion interpretation, and a motion capture system is built to record children's behaviors in the scenario of playing network games. Gunes et al. and Shan et al.~\cite{gunes2007bi,shan2007beyond} propose to extract multi-modality information from videos for emotion recognition, which is more closely related to this paper. Shan et al.~\cite{shan2007beyond} propose to fuse body gestures and facial expressions with Canonical Correlation Analysis (CCA). Furthermore, Gunes et al.~\cite{gunes2007bi} combine features from facial expressions, hand poses, body parts location, and orientation with a late fusion method. Although previous studies have achieved good performances on a bi-modal face and body gesture database (FABO)~\cite{gunes2006bimodal}, there are problems in real applications since the dataset is collected under ideal conditions. FABO contains face and body images in a front view with no body part occlusion, which simplifies the problem. 

\section{Methodology}
\begin{figure*}[t]
\centering
   \centerline{\includegraphics[width=15.8cm]{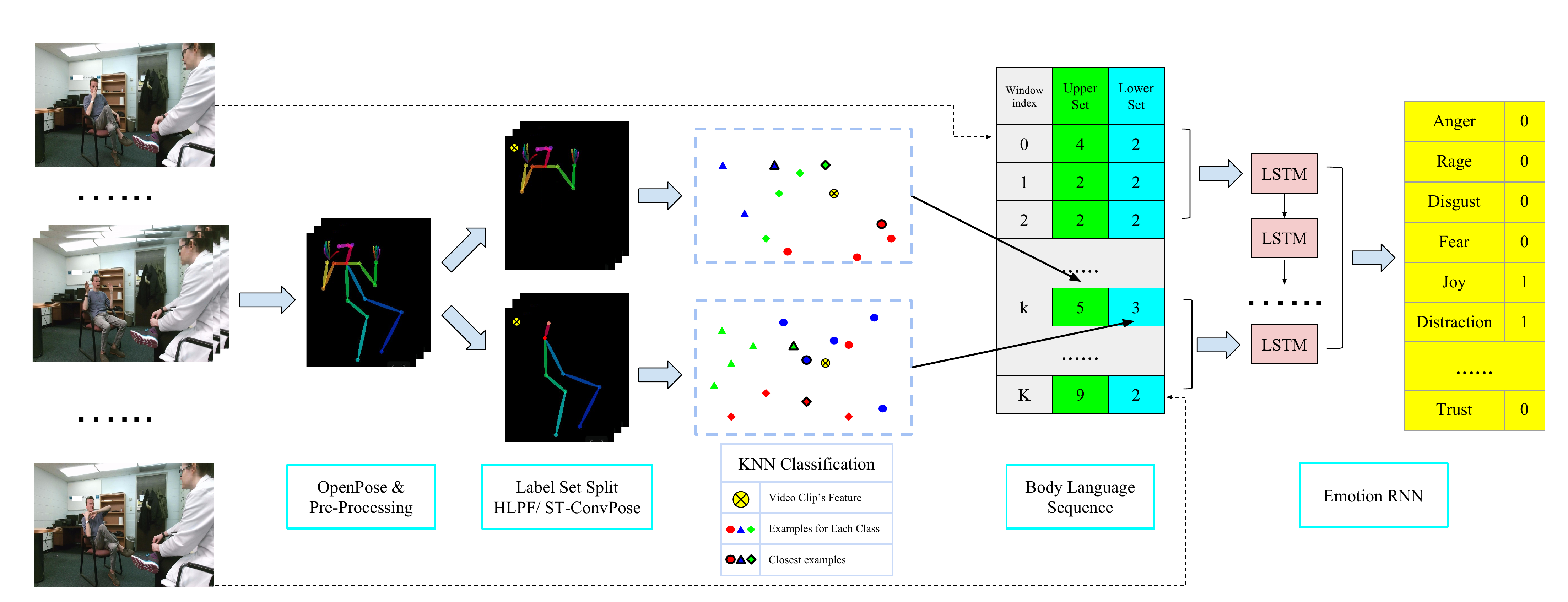}}
	\caption{Overview of the proposed framework. This is a two-stage framework for body language prediction and emotion interpretation. The first stage generates sequences of body language predictions with encoded pose features and a KNN-based classifier. The second stage predicts emotions from the predicted body language sequences.}
\vspace{-0.15in}
\label{fig:framework}
\end{figure*}
\noindent
In this study, we propose a framework that recognizes body language and human emotions with a small amount of data. The input of the framework is a long sequence of estimated human poses $p=\{p_t\}$ where $t \in 0,\ldots,T-1$ is the length of an untrimmed video. The framework contains two stages: the pose-based body language recognition stage and the body language based emotion interpretation stage. In the first stage, the model takes the pose sequence $p$ as the input and outputs two body language sequences that represent the upper- and lower-body body language, respectively. The classifier is built with an easily-transferrable pose feature and an example based classifier, instead of being purely data-driven. The second stage takes the two predicted body language sequences as inputs and learns the emotions expressed by the person of interest. The framework is shown in Figure \ref{fig:framework}.

In this section, we first discuss the pose estimation and pre-processing methods, together with several previous high-level pose feature representations, \ie, the {\it NTraj} and {\it NTraj+}~\cite{jhuang2013towards}.
Then we introduce the proposed Spatial-Temporal Convolutional Pose Feature (ST-ConvPose). We also discuss the methods for processing incorrect or missing pose estimations. After getting the pose feature representation, we predict the body language with an example-based classifier. Finally, we propose a temporal network for emotion interpretation based on the detected body language sequences.

\subsection{Pose Feature Generation}
\label{Pose_feature}
The body language prediction stage uses only the poses as inputs to achieve the desired transferability and to reduce the required amount of data.
We adopt a reliable pose estimator called OpenPose~\cite{cao2016realtime,simon2017hand} to generate 2D human poses from RGB videos. We then encode the estimated poses as the high-level pose features for body language recognition. In this section, we first introduce the pose pre-processing techniques and two previous pose features~\cite{jhuang2013towards}. The proposed pose feature is introduced in the next section.

{\bf Pre-processing.} For each frame, we calculate $18$-point human poses with OpenPose. 
In the pre-processing step, we first normalize the predicted poses with the torso size. In previous works, a puppet mask is used to estimate the torso size, but it is not available in our task. Therefore, we propose to normalize the poses by the distance between the neck joint and the center of two hip joints, such that the length is fixed to $240$ pixels. After the size normalization, all joints are ``centered'' with a reference point. The reference point is defined as the averaged neck joints in several adjacent frames. 

{\bf NTraj.} The {\it NTraj} feature~\cite{jhuang2013towards} is one of the most basic high-level pose features, which only represents the joint position information. Five features are encoded for each joint. The centered x- and y- coordinates are selected as the first two dimensions that represent spatial structures. The differences in pixel locations between two frames are encoded to represent the temporal motion. The direction of motion $arctan(\frac{dy}{dx})$ is also included. Furthermore, the motion is calculated with a temporal gap $s$, i.e. $dx=x_{t+s}-x_t,dy=y_{t+s}-y_t$, which helps eliminate jitters and provide reliable motion. The gap lengths of $s=1,2,3$ are computed, and the trajectory length $T$ is set to $T=5$ based on experiments. Finally, each dimension in the calculated {\it NTraj} features is normalized by the absolute sum value on a training set, i.e. $\frac{(F_i^t,\cdots,F_i^{t+T-1})}{\sum_{j=t}^{t+T-1} \| F_i^j\|}$. The normalization is calculated $5$ times for each feature $F_i^j=\{x^j,y^j,dx^j,dy^j,arctan(\frac{dy}{dx})^j\}$, where $j$ represents the timestamps.

{\bf NTraj+.} Besides the 5 kinds of position-based features in {\it NTraj} calculated at each joint independently, the {\it NTraj+} feature further includes the relationship information among different joints. The orientations between every two joints and the inner angles of all permutations in a three-joint group are encoded to represent the relationship information.

The bag-of-features is calculated to encode the pose feature representations. For each feature type, a codebook of size $N$ is formed by running k-means $10$ times on all features available in a training set and pick the one with the lowest error. With a small feature dimensionality, the codebook size $N$ is tested among $N= 10,20,50,100,200,500$. 
\subsection{ST-ConvPose Feature}
\begin{figure}[t]
\centering
	\includegraphics[width=8cm]{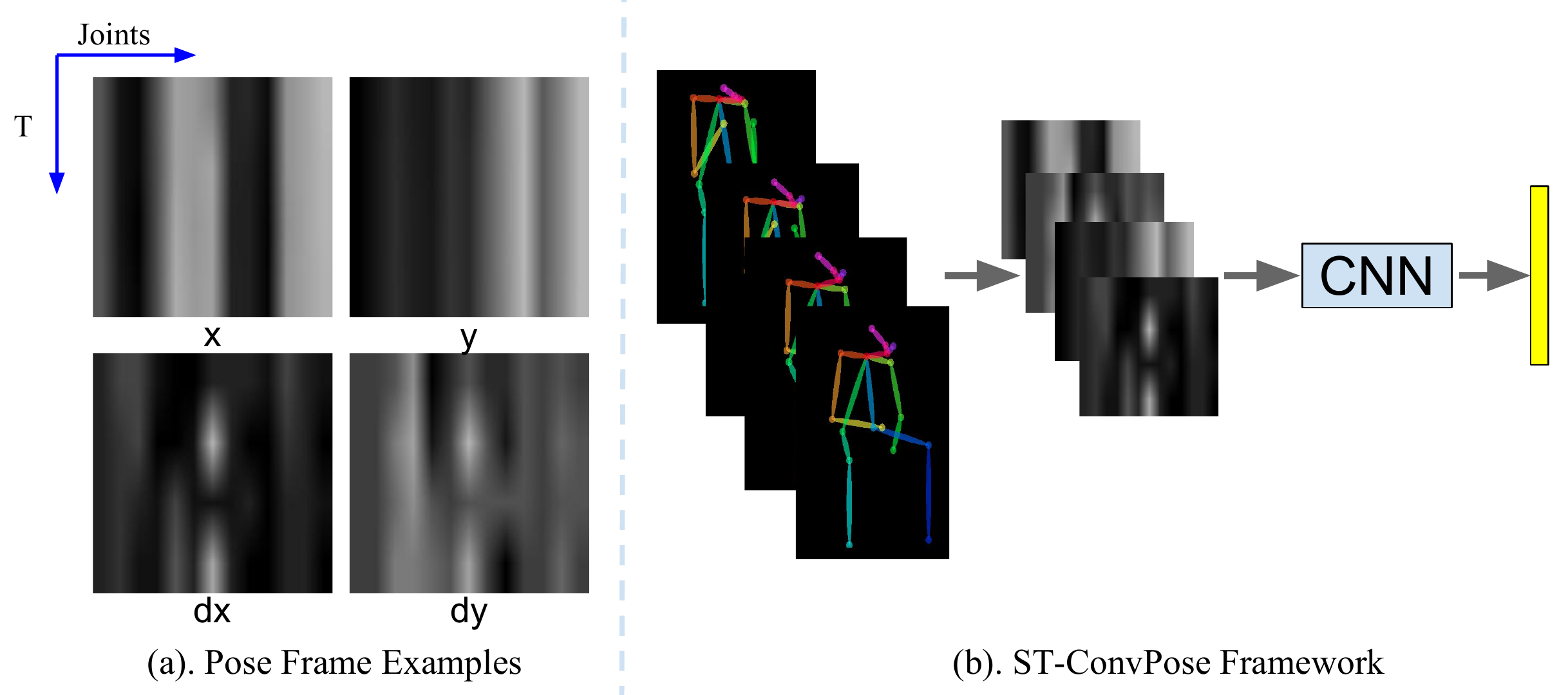}
	\caption{(a). Pose image examples. (b). The framework for the spatio-temporal convolutional pose feature (ST-ConvPose).}
\vspace{-0.1in}
\label{fig:stconvpose}
\end{figure}

In previous high-level pose features, the spatial relations among joints are encoded with pre-defined features such as body part lengths, orientations, inner angles, and so on. Many RGB+Pose action recognition studies~\cite{cheron2015p,cao2016action} also only use the pre-defined spatial relation information among joints that can not be adaptively learned.
Inspired by the recent studies on 3D skeleton representations~\cite{ke2017new,kim2017interpretable,wang2017modeling}, we propose a spatial-temporal convolutional pose feature (ST-ConvPose) that learns the spatial and temporal relations among joints simultaneously with 2D CNNs. In the proposed feature, the coordinates of poses are arranged as 2D matrices where each row is the
chaining of joint coordinates at time-stamp $t$, and the column lists the chains in all timestamps. 
To arrange the joints in each row, we experiment with three different orders proposed by J-HMDB~\cite{jhuang2013towards}, PennAction~\cite{zhang2013actemes}, and NTU RGB+D~\cite{shahroudy2016ntu}. The order defined in NTU RGB+D works the best on action recognition datasets. Therefore, we arrange the joints in each row following this order that joints in each body part are grouped first and then chained from upper left to lower right. To be specific, the left shoulder to the left wrist are placed in column 2 to 4, the right shoulder to the right wrist are in column 5 to 7, and so on. The encoded 2D matrices are then scaled into 0 to 255 and resized to a fixed width and height. The formatted multi-channel 2D matrices are referred to as pose images. Examples of the generated pose images are shown in Figure \ref{fig:stconvpose} (a). The pose images are then fed into CNNs to generate human activity representation in an end-to-end manner. The ResNet-50~\cite{he2016deep} is used as the CNN structure for high-level pose feature encoding. The feature encoding structure is shown in Figure \ref{fig:stconvpose} (b). 

With the ability to jointly learn the spatio-temporal joint relations, the proposed ST-ConvPose features show a better performance in modeling human activities compared to previous high-level pose features. Furthermore, the ST-ConvPose feature can be trained with a limited amount of training data and possesses a good transferability. With these desired properties provided by pose features, we do not include RGB frames as inputs to the framework and learn purely based on estimated poses. More details are discussed in the experiment section.

\subsection{Body Language Sequence Prediction}
\begin{figure*}[t]
\centering
	\includegraphics[width=16cm]{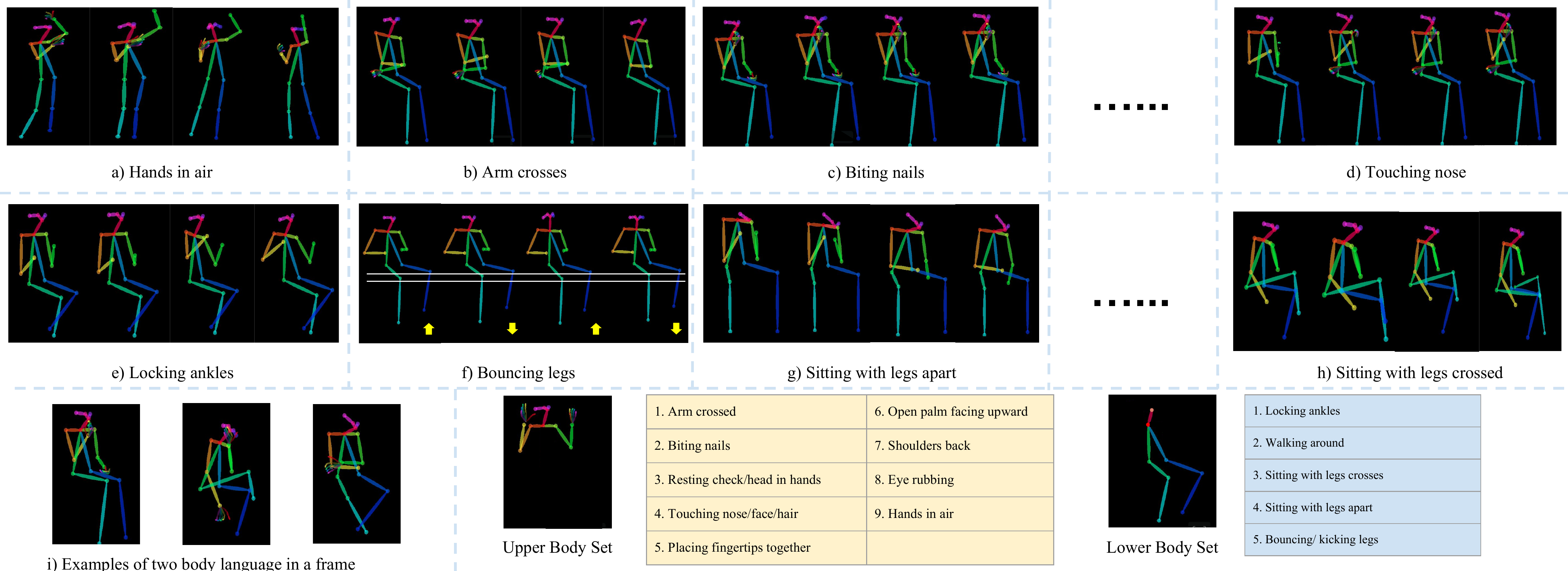}
	\caption{Examples from the URMC dataset. First row:  examples from the upper body set. Second row: examples from the lower body set. Third row: examples of two body language appearing simultaneously. The complete label sets are also included.}
\label{fig:example}
\end{figure*}

The proposed body language recognition stage is designed to work on untrimmed videos, and a fixed-length sliding window is adopted to convert the localization and recognition problem into a regular classification task. 

In practice, different body parts could perform different body language simultaneously, and the required joints for recognition are different. For example, people might keep a catapult posture while having their legs crossed. Because of this, we divide body language into two label sets and predict them separately. The lower body pose set contains leg postures, including locking ankles, leg crossed, walking around, and so on. The upper body pose set includes both arm motions and hand movements. A background class is added to both pose sets. Full label name list is shown in Figure \ref{fig:example}. The split of the labeling set not only solves the multi-label problem but also provides classifiers the prior knowledge on which groups of joints should be focused on to recognize a specific body language. Joints on the torso and both legs are used for lower-body body language recognition, and the upper body recognition task takes the joints on the arms and head as input.

After obtaining the feature representations with either {\it NTraj+} or ST-ConvPose, we adopt a k-nearest neighbor classifier with manually selected label examples. For each body language, we manually select five to seven video clips and use their pose features as the KNN data points. Classification is conducted on each video clip under the sliding window. The output of the model is the two sequences of upper and lower body language predictions. The body language prediction framework is shown in the left part of Figure \ref{fig:framework}. 

\subsection{Emotion Interpretation}
After getting the body language sequence predictions, we propose an emotion RNN network for emotion interpretation as the second stage of the framework. We predict the body language sequences for all video clips in a dataset. Each output contains the lower- and upper-body body language prediction sequences with length $K$, where $K$ is the number of sliding windows in an input video. We apply another sliding window on the predicted body language sequences and calculate the body language histogram under each window. The histogram sequences are fed into an LSTM network in the temporal order. The LSTM outputs at all timestamps are fed into another dense layer for the final video-level emotion prediction. The output is an N-hot vector representing the predicted emotions.

We have also experimented with the end-to-end emotion recognition model. We replace the KNN classifier with dense layers and teach the model to simultaneously predict body language and emotions. Unsurprisingly, due to the limited size of the available emotion labels, the end-to-end framework fails to learn effectively.

\section{The URMC Dataset}
Among the various applications of the proposed body language and emotion recognition framework, psychotherapy is an area especially suitable for model evaluation. There exist solid and complete theoretical proofs in the medical field for the emotions of interest and the corresponding body language. Therefore, the proposed framework is employed to predict psychiatric symptoms. We construct a new URMC dataset under realistic simulated mental health care scenarios with expert annotations. A typical scenario involves the conversation between a standardized patient and a psychiatrist, with an initial intention to infer the standardized patient's potential symptoms by analyzing their body language. There are $144$ 30-second long video segments cropped from $12$ 20-minute videos.

\subsection{Dataset Collection}
\label{41}
{\bf Scene Recording.} We collaborated with psychologists to collect 12 videos recording the diagnostic sessions between standardized patients (highly trained actors) and mental health professionals, with the view focused on the standardized patients (SPs). The mental health professionals complete a survey after each diagnostic session to control the quality of the dataset. Positive feedbacks on questions, including the confidence level of diagnosis and other session-related questions, prove the quality of the recordings. All participants in the recordings provided permission for the data collection for research. Both RGB videos and conversation audios are recorded, and poses are later estimated using RGB videos. Example frames and estimated poses are shown in Figure \ref{fig:framework}. Four standardized patients and eight mental health professionals (psychologists, psychiatrists)  are involved. For the 12 videos, either the standardized patient or the psychiatrist is different. Also, the subjects' clothing is different in all videos.

{\bf Dataset Labeling.} Three mental health professionals are involved in the dataset labeling process. In the first step, three mental health professionals go through each of the 12 recorded long videos, remove less informative segments, select 12 30-second video segments, and then make the symptom decisions. Removing the less informative segments can reduce the labeling cost, and also reduce the noises in the dataset.
After getting the 144 video segments, three mental health experts label the video clips with discernible body language and symptoms. Each video clip is labeled by all three health professionals, and conflicts are resolved in consensus discussions. Both the video recordings and the conversation audios are used for labeling, as certain symptoms can only be reflected from the content of the conversation. It is worth mentioning that instead of generating body language and emotion labels based on previous studies in the computer science field, the labels for annotation in our dataset are designed by the mental health experts with evidence in the medical field. Based on the psychiatry theories, 32 body language, 24 emotions, and 24 symptoms are selected for labeling. A final psychiatric verdict of whether the standardized patient has the major depressive disorder (MDD) or manic episode (ME) is also included. This dataset is built to help inferring emotions and psychiatric symptoms from body language.

The 32 body language includes {\it arm crossed, finger tapping, ear pulling and etc.}  Once a body language appears in a 30-second video, the video clip is marked with that label. Because certain body language rarely appear and several of them are extremely similar, we merge a number of body language classes. The final 14 body language classes are shown in Figure \ref{fig:example}, which are {\it arm crossed, biting nails, and etc}.
Furthermore, a subset of the videos clips has frame-level body language labels that are annotated under sliding windows with a length of 6 frames. 


\subsection{Dataset Summary}
There are $144$ 30-second video clips cropped from $12$ 20-minute videos. Each clip has multiple labels from the 14 body language, 24+1 emotion labels, and 24+1 symptom labels. Among the 144 videos, 48 videos are selected for training and have  frame-level body language labels. Additionally, 48 videos are for validation, and 48 videos are for testing. The split follows a cross-subject setting. We will release the collected and processed poses.

\section{Experiments}
In this section, we first experiment with the proposed ST-ConvPose feature. We then evaluate the proposed body language and emotion recognition framework on the URMC dataset. Section \ref{53} presents the results of body language prediction, and Section \ref{54} shows the performance of emotion recognition and psychiatric symptom prediction.

\subsection{ST-ConvPose Feature Evaluation}
\label{51}

Because there lack public benchmarks on body language recognition, we evaluate the proposed ST-ConvPose feature on a similar task of action recognition and compare our method to state-of-the-art methods. It is important to note that although body language, as defined earlier, {\it are different from actions}, it is the best comparison we could perform due to the lack of existing body language datasets. Experiments prove that the proposed ST-ConvPose feature outperforms both RGB-based and pose-based state-of-the-art with only the pose information. Furthermore, experiments show that the ST-ConvPose feature achieves the following two desired properties:
\begin{enumerate}
\item The pose feature generates good results when the amount of training data is limited. 
\item The pose feature captures more dataset-invariant human representations instead of learning the dataset bias, thus has a better transferability.
\end{enumerate}
We detail the analyses as follows.

\vspace{3pt}
\noindent{\bf Action Recognition Results.} 
We evaluate the proposed ST-ConvPose feature on the PennAction~\cite{zhang2013actemes} and UCF-Motion~\cite{yang2018action} datasets. PennAction contains 1,212 videos for training and 1,020 videos for testing, with 2D full body joints manually labeled on each frame. Half of the training videos are used for training, and the rest is the validation set. UCF-Motion dataset extends UCF-101 \cite{soomro2012ucf101} by including estimated poses~\cite{fang2017rmpe} on all video frames. UCF-Motion contains 23 classes from UCF-101 with 3172 videos in total.
Table~\ref{table:pennaction} shows the action recognition accuracy on the PennAction dataset. 
Our ST-ConvPose feature outperforms both the RGB-based and the pose-based state-of-the-art on the PennAction dataset. It is gratifying even when compared with the methods using both RGB and pose information, the proposed pose feature can achieve comparable results. The experiments show the effectiveness of the proposed ST-ConvPose feature and prove that poses can adequately represent the information needed to distinguish human actions. A similar improvement is also observed on UCF-Motion, as shown in Table~\ref{table:UCF}.

\begin{table}[t]
\centering
\caption{Recognition accuracy compared to the state-of-the-art on PennAction. The input data format is also shown.} 
\begin{tabular}{ c c c c }
    \hline
    State-of-the-art & Acc. & Pose & RGB\\
    \hline
	C3D \cite{tran2015learning} & 86.0 & - & \checkmark\\
	idt-fv \cite{iqbal2017pose} & 92.0 & - & \checkmark \\
	NTraj+ \cite{iqbal2017pose} & 79.0 & \checkmark & -\\
	JDD \cite{cao2016action} & 87.4 & \checkmark & \checkmark\\
	Pose+idt-fv \cite{iqbal2017pose} & 92.9 & \checkmark & \checkmark\\
    \hline
	ST-ConvPose & {\bf 94.4} & \checkmark & -\\
\end{tabular}
\label{table:pennaction}
\end{table}
\begin{table}[t]
\caption{The action recognition accuracy compared to the state-of-the-art methods on the UCF-Motion dataset.}
\centering
\begin{tabular}{ c c c c c}
    \hline
    State-of-the-art & Acc. & Pose & RGB & Flow\\
    \hline
	HLPF \cite{jhuang2013towards} & 71.4 & \checkmark & - & - \\
	C3D \cite{tran2015learning} & 75.2 & - & \checkmark & - \\
	Flow CNN \cite{kay2017kinetics} & 85.1 & - & - & \checkmark \\
    \hline
	ST-ConvPose & {\bf 88.1} & \checkmark & - & - \\
\end{tabular}
\label{table:UCF}
\end{table}
\begin{table}[t]
\centering
\caption{Experiments on training with less data. $100\%, 50\%, 20\%$ are the percentage of used training data. {\it ``NTU Pretrain''} indicates that the ST-ConvPose feature is pretrained on NTU RGB+D. {\it ``Drop''} is the accuracy decrease percentage when using $20\%$ training data compared to using all data.}
\begin{tabular}{ c c c c c c c }
    \hline
    State-of-the-art & $100\%$ & $50\%$ & $20\%$ & $Drop\%$\\
    \hline
	C3D \cite{tran2015learning} & 86.0 & 74.1 & 65.0 & 24.4 \\
    \hline
    ST-ConvPose & $100\%$ & $50\%$ & $20\%$ & $Drop\%$\\
    \hline  
    ST-ConvPose from Scratch & 94.4 & 84.3 & 79.6 & 15.7 \\
	NTU Pretrain. w/o Fine-tune & 82.4 & 77.6 & 70.0 & 15.0 \\
	NTU Pretrain. with Fine-tune & 95.4 & 93.2 & 90.4 & {\bf 5.2} \\
\end{tabular}
\label{table:lessdata}
\end{table}

\vspace{3pt}
\noindent{\bf Training with Less Data.} Our method also performs well with a limited amount of training data. We design the experiments by sampling different amounts of training data from PennAction. Starting from using all the training data, we train the model with $50\%$ and $20\%$ of the training set. All 1,020 testing videos are used for testing. The training and validation sets are still split. As shown in Table \ref{table:lessdata}, when $20\%$ of the training data is used, the performance of RGB-based methods drop $(86.0\%-65.0\%)/86.0\%=24.4\%$, while ST-Convpose only drops $15.7\%$. The performance drop is even smaller when ST-Convpose is pretrained on NTU RGB+D~\cite{shahroudy2016ntu} of $5.2\%$. This experiment shows that the ST-ConvPose works well with a limited amount of training data.

\vspace{3pt}
\noindent{\bf Domain Transfer.} 
Furthermore, pose feature captures dataset-invariant action representations and thus has a better ability to transfer across datasets, compared to RGB-based methods. We adopt the ST-ConvPose feature trained on NTU RGB+D with projected x and y coordinates for action recognition on the PennAction dataset with or without fine-tuning.

As shown in Table~\ref{table:lessdata}, the ST-ConvPose feature achieves an $82.4\%$ accuracy using the pose features pretrained on NTU RGB+D without fine-tuning, which is already better than the $79.0\%$ accuracy generated by previous $NTraj+$ pose feature. The high recognition accuracy proves that the pose feature captures invariant information to represent actions and has an excellent ability for domain transfer even without fine-tuning. The results can be further improved with fine-tuning. 

In this experiment, we show that the proposed ST-ConvPose feature outperforms the state-of-the-art action recognition methods on PennAction and UCF-Motion. Furthermore, it requires less training data and has good transferability.  

\subsection{Evaluation Settings on the URMC Dataset}
We evaluate the proposed body language and emotion recognition framework on the URMC dataset. For the first stage of the framework, we treat the body language sequence prediction as a video-level multi-label classification task. In the experiment, an N-hot vector is generated from the predicted body language sequence and compared with the video-level ground truth body language labels. Metrics for multi-label classification are used for evaluation, \ie, multi-label accuracy, precision, recall, and F1-score. 

The second stage of the framework is emotion interpretation and symptom prediction. For emotion interpretation, we compare the proposed temporal network with other approaches as a multi-label classification task. Instead of predicting all 24 labeled psychiatric symptoms, the model only learns to distinguish major symptoms of Major Depressive Disorder (MDD) with Manic Episode (ME). As mentioned in Section \ref{41}, certain symptoms can only be reflected with other modalities, such as the audio track. Although predicting detailed psychiatric symptoms is a very interesting problem of great importance, we focus on a more feasible task of inferring emotions and major psychiatric symptoms in this study.

\subsection{Body Language Recognition}
\label{53}
\begin{table*}
\centering
\caption{Multi-label body language recognition performance compared with different approaches.}
\begin{tabular}{ c c c c c c c c c}
    \hline
    Lower Body Set & Acc. & Prec. & Recall & F1 & Interpretability & transferability & Required Data Size \\
    \hline
	Two Stream & 0.445 & {\bf 0.581} & 0.497 & 0.526 & & & Large \\
	$NTraj^+$+SVM & 0.327 & 0.336 & {\bf 0.690} & 0.424 & & & Medium\\
	$NTraj^+$+KNN & 0.483 & 0.516 & 0.616 & 0.538 & \checkmark & & Small\\
	ST-ConvPose+Dense & 0.384 & 0.397 & 0.606 & 0.460 & & \checkmark & Medium\\
	{\bf ST-ConvPose+KNN} & {\bf 0.488} & 0.520 & 0.658 & {\bf 0.554} & \checkmark & \checkmark & Small\\
    \hline
    Upper Body Set & Acc. & Prec. & Recall & F1 & Interpretability & transferability & Required Data Size \\
    \hline
	Two Stream & 0.341 & 0.472 & 0.567 & 0.492 & & & Large\\
	$NTraj^+$+SVM & 0.346 & 0.388 & {\bf 0.766} & 0.485 & & & Medium\\
	$NTraj^+$+KNN  & 0.398 & 0.504 & 0.578 & 0.502 & \checkmark & & Small\\
	ST-ConvPose+Dense & 0.374 & {\bf 0.522} & 0.486 & 0.473 & & \checkmark & Medium\\
	{\bf ST-ConvPose+KNN} & {\bf 0.400} & 0.497 & 0.641 & {\bf 0.519} & \checkmark & \checkmark & Small\\
    \hline
\end{tabular}
\label{table:multi_baseline}
\end{table*}

Table~\ref{table:multi_baseline} shows the body language recognition accuracy. Other than the accuracy metrics, we also indicate the models' interpretability, transferability, and the required amount of training data. We first compare the model with two end-to-end methods: the two-stream action recognition and the proposed ST-ConvPose feature trained end-to-end with dense layers. The two-stream action recognition is pretrained on UCF101~\cite{soomro2012ucf101} and HMDB51~\cite{kuehne2011hmdb}. Since both methods are trained end-to-end, they lack the interpretability. The ST-ConvPose feature requires less data since it is trained from pose information instead of RGB frames. Furthermore, the ST-ConvPose feature provides a good ability for domain transfer. For SVM, we adopt the one-vs-one approach for multi-class classification with linear kernels. Overall, the KNN classifier provides interpretable results and requires less data for training. The proposed ST-ConvPose feature provides the desired transferability and further reduces the required amount of training data.

In Table~\ref{table:multi_baseline}, $NTraj^+$ + SVM and ST-ConvPose + Dense layers have the lowest recognition accuracy because of the limited training data size of 48 video clips. Although the two-stream network requires even more training data, the model is transferred from the one pretrained on UCF101 and HMDB51. In contrast, our KNN based methods have the best performance with a small training set. Overall, our proposed ST-ConvPose and the KNN classifier has the best recognition performance.

\subsection{Emotion and Symptom Prediction}
\label{54}
\begin{table}
\centering 
\caption{The performance of emotion interpretation by  different learning methods and body language sequences. $L$ is the length of the sliding window and $S$ is the stride of the window.} 
\begin{tabular}{ c c c c c c }
    \hline
    LSTM+ST-ConvPose & Acc. & Prec. & Recall & F1 \\
    \hline
	$L=1$,\ $S=1$ & 0.468 & 0.644 & 0.630 & 0.637 \\
	$L=7$,\ $S=3$ & {\bf 0.564} & 0.775 & {\bf 0.674} & {\bf 0.721} \\
	$L=48$ & 0.145 & 0.162 & 0.587 & 0.254\\
    \hline
    Other Methods & Acc. & Prec. & Recall & F1 \\
    \hline
	LSTM+$NTraj^+$ & 0.510 & {\bf 0.839} & 0.565 & 0.675 \\
	Conv 1D+$NTraj^+$ & 0.490 & 0.788 & 0.565 & 0.658 \\
	Conv 1D+ST-ConvPose & 0.556 & 0.789 & 0.652 & 0.714 \\
    \hline
\end{tabular}
\label{table:emotion}
\end{table}
The ultimate goal of the framework is to understand emotions and infer psychiatric symptoms from human body language. We conduct emotion recognition purely based on the predicted body language sequence to achieve the desired transferability. A LSTM-based temporal network is proposed for the task. The output of the network is an N-hot vector with a length of 25, representing the 24 labeled emotions and a background class. For baseline comparisons, we implement a network with a 1D convolution for temporal information learning. The idea of adopting a convolutional layer for sequence learning is inspired by works in natural language processing, where CNN are used for sentence analysis~\cite{kim2014convolutional,kalchbrenner2014convolutional}. The body language sequences predicted by the proposed ST-ConvPose feature is also compared with the one predicted with {\it NTraj+}. We also consider an end-to-end design of the framework. However, the result is not promising due to the limited amount of body language labels and emotion labels. 

The experiment results on emotion interpretation are shown in Table \ref{table:emotion}.
Two special cases are worth noting. First, when the sliding window length and stride both equal to one, the predicted vector sequences are directly fed into the temporal network without forming a histogram. Second, when the window length is the entire video, a video-level body language histogram is calculated with no temporal information used, similar to directly adopting a dense layer at the top. The limited performance of the two special cases proves the effectiveness of the proposed temporal structure and histogram features. The experiment results in Table \ref{table:emotion} also indicate that the LSTM structure has a better ability in representing the temporal information in body language sequences, compared to 1D convolutional neural networks. Additionally, using the proposed ST-ConvPose feature generates a better overall performance compared to previous high-level pose features.

Furthermore, we try to infer the psychiatric symptoms based on the body language predictions. The result of the binary classification between Major Depressive Disorder (MDD) or Manic Episode (ME) is promising. We achieve an accuracy of $90.3\%$ with the ground truth body language sequences and $79.9\%$ with the predicted sequences. 

\section{Conclusion and Future Work}
Teaching machines to recognize human emotions from body language is a challenging but useful task that has various potential applications. In this paper, we propose a framework starting from regular  RGB videos for body language prediction and emotion interpretation. Aiming at building a system capable of easy transfer between different application scenarios and producing interpretable results, we design a body language recognition system with the ST-ConvPose feature and the KNN classifier. The proposed pose feature shows good performances on both public datasets and the new URMC dataset, while requiring less training data and showing good transferability. Finally, we show that emotions or psychiatric symptoms can be predicted reliably from recognized human body language. Next on our agenda is to apply the framework to more domains and continue exploring detailed psychiatric symptom detection based on body language.

\section{Acknowledgement}
This work is supported in part by NSF awards IIS-1704337, IIS-1722847, and IIS-1813709, Twitch Fellowship, a grant from the Goergen Institute for Data Science Collaborative Pilot Award Program in Health Analytics, and the Morris K.Udall Center of Excellence in Parkinson's Disease Research by NIH.


\bibliographystyle{IEEEtran}
\bibliography{refs}

\end{document}